\documentclass[letterpaper, 10 pt, conference]{ieeeconf}  
\usepackage{times,latexsym,macros}
\usepackage{url}
\usepackage[T1]{fontenc}
\usepackage{hyperref}
\usepackage{amssymb}
\usepackage{amsmath}
\usepackage{cleveref}
\usepackage{listings}
\usepackage[dvipsnames]{xcolor}
\usepackage{subcaption}
\expandafter\def\csname ver@subfig.sty\endcsname{}
\usepackage{graphicx}
\usepackage{booktabs}
\usepackage{subfig}
\usepackage{multirow}
\usepackage{tikz}
\definecolor{codegreen}{rgb}{0,0.6,0}
\definecolor{codered}{rgb}{0.8,0,0}
\definecolor{codeblue}{rgb}{0.0,0,0.6}
\definecolor{codegray}{rgb}{0.5,0.5,0.5}
\definecolor{codepurple}{rgb}{0.58,0,0.82}
\definecolor{backcolour}{rgb}{0.95,0.95,0.92}
\usepackage{pgfplots}
\usepackage{xurl}

\newcommand\rurl[1]{%
  \href{http://#1}{\color{magenta}\nolinkurl{#1}}%
}

\lstdefinestyle{mystyle}{
    commentstyle=\color{codegreen},
    keywordstyle=\color{black},
    numberstyle=\tiny\color{codegray},
    stringstyle=\color{codeblue},
    basicstyle=\ttfamily\footnotesize\color{black},
    breakatwhitespace=false,         
    breaklines=true,                 
    captionpos=b,                    
    keepspaces=true,                 
    numbers=left,                    
    numbersep=5pt,                  
    showspaces=false,                
    showstringspaces=false,
    showtabs=false,                  
    tabsize=2,
    frame=single,
    xleftmargin=1.5em,
    framexleftmargin=1.5em
}
\usepackage{cleveref}
\crefname{section}{\S}{\S\S}
\Crefname{section}{\S}{\S\S}
\crefname{table}{Table}{Tables}
\crefname{figure}{Figure}{Figures}
\crefname{algorithm}{Algorithm}{}
\crefname{equation}{eq.}{}
\crefname{appendix}{App.}{}
\crefformat{section}{\S#2#1#3}

\usepackage{amsmath}               
  {

  }

\IEEEoverridecommandlockouts                              

\title{\LARGE \bf
NARRATE: Versatile Language Architecture for Optimal Control in Robotics
}

\author{Seif Ismail$^*$, Antonio Arbues$^*$, Ryan Cotterell, René Zurbrügg, Carmen Amo Alonso
\thanks{
$^*$Equal Contribution. The authors are with ETH Zürich. This research was partially supported by the ETH AI Center.\ }
}
\pgfplotsset{compat=1.18} 

\begin{document}

\maketitle

\thispagestyle{empty}
\pagestyle{empty}

\begin{abstract}

The impressive capabilities of Large Language Models (LLMs) have led to various efforts in enabling robots to be controlled through natural language instructions, opening exciting possibilities for human-robot interaction.
The goal is for the motor-control task to be performed accurately, efficiently and safely while also enjoying the flexibility imparted by LLMs to specify and adjust the task through natural language. 
In this work, we demonstrate how a careful layering of an LLM in combination with a Model Predictive Control (MPC) formulation allows for accurate and flexible robotic control via natural language while taking into consideration safety constraints.
In particular, we rely on the LLM to effectively frame constraints and objective functions as mathematical expressions, which are later used in the motor-control module via MPC. 
The transparency of the optimization formulation allows for interpretability of the task and enables adjustments through human feedback. We demonstrate the validity of our method through extensive experiments on long-horizon reasoning, contact-rich, and multi-object interaction tasks. 
Our evaluations show that \ourmethod outperforms current existing methods on these benchmarks and effectively transfers to the real world on two different embodiments.\\ Videos, Code and Prompts at \rurl{narrate-mpc.github.io}

\end{abstract}

\section{INTRODUCTION}

Efficiently translating high-level language instructions into precise low-level actions for robots and autonomous agents presents an important challenge in both artificial intelligence and robotics.  
While today's robots excel at following precise, low-level instructions (such as joint commands, end-effector poses, or cartesian trajectories), most systems require them to be pre-programmed for each specific task. 
This limits the potential for seamless integration of robots into daily life or their application to diverse and complex tasks. 
Natural language instructions offer a solution, as they allow humans to describe tasks in a familiar and flexible way without the need for intricate and expert programming.

\begin{figure}[t]
\centering
\includegraphics[width=\linewidth,trim={0 2mm 0 0},clip]{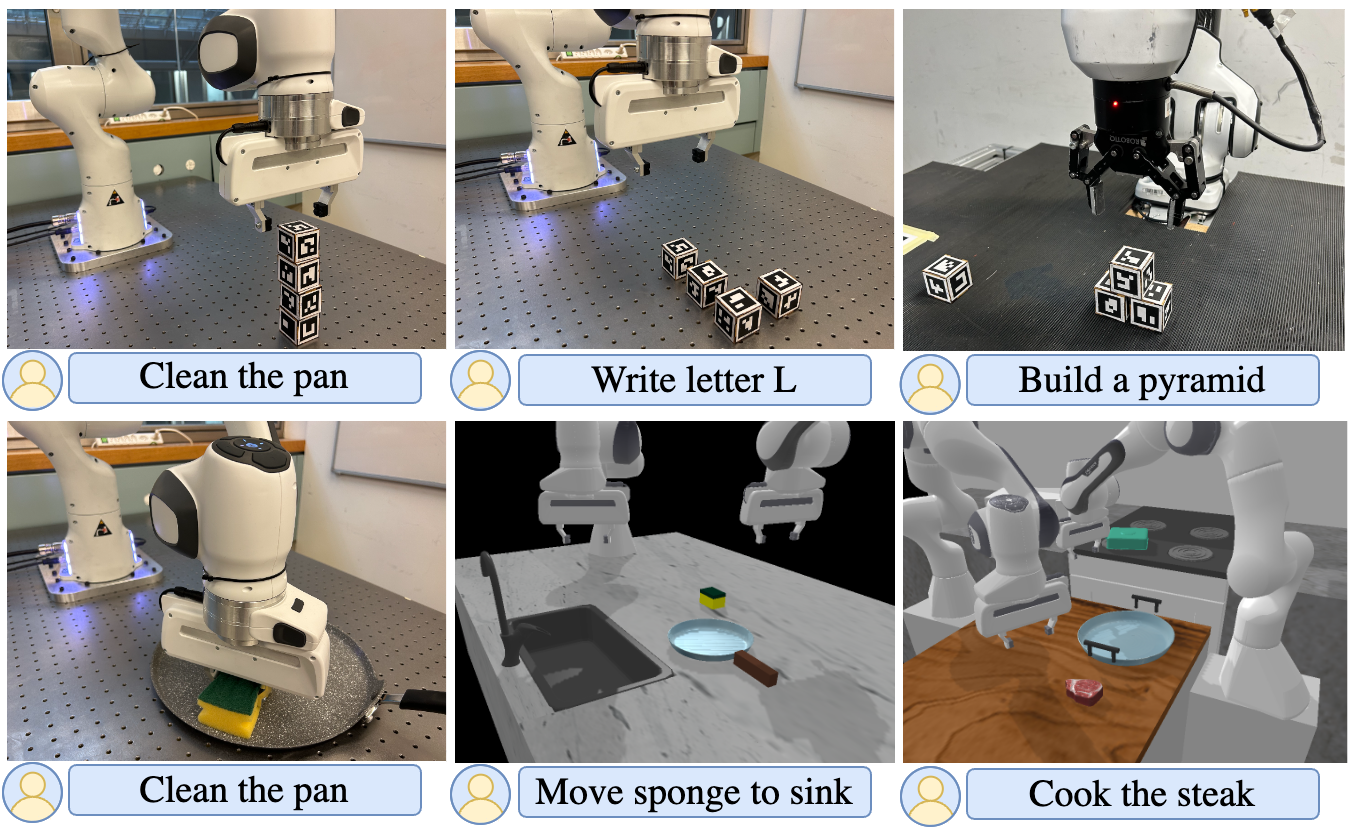}
    \caption{\textbf{Benchmark Tasks:} Given an instruction in natural language, the robotic manipulator is able to autonomously perform a variety of different tasks, both in simulation and on different real robots.}
    \label{fig: all_tasks} \vspace{-12px}
\end{figure}

Large language models (LLMs) have exhibited potential beyond natural language  \cite{hausman2023natural, shah2023navigation, lake2023human}.
Their ability to extract semantics and generate plans according to the context of their instructions makes them well-suited for control from natural language, which is often underspecified and relies heavily on the given context. 
Yet, grounding their output in low-level actions remains an active area of research \cite{singh2023progprompt,vemprala2023chatgpt}. 
As of today, a large number of methods aim to directly predict actions conditioned on language and visual observations using self-supervised and imitation learning \cite{2307.15818, open_x_embodiment_rt_x_2023, 2306.11706, 2401.02117}. While these methods have shown impressive results, they are bottlenecked by the limited amount of robot- and task-specific data available. Moreover, their end-to-end architecture hinders interpretability and does not allow for insights into their decision-making.

Recent work has tried to overcome the limited availability of robotics training data by using generative models to synthesize artificial datasets \cite{2311.01455} or by asking pretrained LLMs to directly output low-level actions \cite{ 2309.09969}. Another commonly adopted alternative utilizes LLMs to output predefined expressions, such as motion- \cite{liang2023code} or reward-function primitives \cite{yu2023language}. However, restricting the network's output to the form of primitives generally limits it to a fixed library of skills, losing some of the adaptability that was gained from the introduction of LLMs.

To make full use of the versatility of LLMs, we introduce \ourmethod (Natural language Architecture for Robot Control and Adaptive Task Execution), a modular architecture that interfaces language models with optimization-based controllers. 
Our main contribution is the use of LLMs to predict free-form mathematical expressions that are directly used with Model Predictive Control (MPC) \cite{rawlings2017model} to accurately solve manipulation tasks specified in natural language. This allows us to overcome limitations related to using pre-defined primitives while still not needing robot control data. The simplicity of this method allows for writing general formulations, such as time-dependent functions, and the seamless addition of constraints into the policy.
To the best of our knowledge, the incorporation of hard constraints into robotic policies prompted by LLMs has not been explored before, and, as we show in this work, it significantly improves the performance and efficiency of the closed-loop controller. In this way, our approach takes full advantage of the flexibility that LLMs offer while also ensuring accuracy and providing higher safety during task performance. Moreover, the relatively low latency required for LLM inference and the real-time properties of MPC allow for a very fast and intuitive interface between human users and NARRATE. This accommodates the inclusion of human feedback in natural language which can sensibly increase the system's adaptability and performance.  

Our system is evaluated on a wide variety of tasks, ranging from long-horizon planning to multi-robot interactions. We further deploy \ourmethod on two distinct robot manipulators and qualitatively evaluate our method on one of them, showcasing how effectively and seamlessly it transfers to the real world across multiple embodiments. In summary, our contributions are:
\begin{enumerate}
    \item We introduce a new method to formulate general MPC policies from high-level natural language instructions for complex manipulation tasks. 
    \item We extensively evaluate \ourmethod in simulation and hardware across different embodiments.
    \item We show how \ourmethod accommodates the natural interaction between humans and LLMs and how this collaboration improves the system performance.
\end{enumerate}

\section{Related Work}

In \defn{language-to-action} approaches, language instructions are directly mapped to low-level control actions \cite{jiang2022vima,wu2023visual,Shridhar2022PerceiverActorAM}. 
In order to ensure good performance, this approach requires a significant amount of training data, either as offline datasets \cite{jiang2022vima} or online environment interaction \cite{wu2023visual,Shridhar2022PerceiverActorAM}. 
Furthermore, the scalability of this approach to robot collaboration settings is limited.
For instance, Zhao et~al. \cite{mandi2023roco} require extensive \emph{a posteriori} sanity checks and often recomputation in order to find a suitable trajectory that avoids collisions among agents, which significantly hinders the efficiency of this approach.\looseness=-1

In a \defn{language-to-code} approach, language instructions are mapped into executable code \cite{austin2021program}. 
This approach capitalizes on the ability of LLMs to write code that better captures the expressivity of natural language in a few-shot manner without specific training. It has proven successful for a variety of tasks \cite{li2022competition,trivedi2021learning}. For instance, in the \textit{Code-as-Policies} approach \cite{liang2023code}, an LLM is used to receive instructions in natural language and produce Python code encoding robot policies. This idea has further been tested in AI-embodied agents for navigation in virtual environments   \cite{wang2023voyager}. However, the effectiveness of these techniques is often hindered by the fact that code generally provides a very unstructured setting for robot control, together with the limitations that LLMs exhibit at reliable code writing \cite{openai2023gpt4}, which can lead to consistency or even safety issues. To overcome this limitation, these approaches often use human-engineered control primitives, which in turn heavily restrict the flexibility of capturing nuances in diverse settings. 

In order to overcome these limitations, \defn{language-to-reward} approaches have been proposed \cite{goyal2019using}. In this framework, domain-specific models that map language instructions to reward functions are used. Recent work also tries to extract implicit reward functions via specific representations, such as 3D composable maps, that can be generated with code written by pre-trained LLMs \cite{wang2023voyager}. 
An advantage of this approach is that these rewards can be used in optimization-based controllers, such as Model Predictive Control, to generate optimal control trajectories \cite{yu2023language}. Since human instructions can often be formulated as an objective function to be optimized under some conditions (constraints), optimal control problems seem to be a good interface for translating human language into robot actions \cite{arora2021survey}.
However, when only rewards are modeled, the space of possible actions has to be restricted in order to ensure safety and maintain acceptable success rates (accuracy-robustness tradeoff). In particular, the work in \cite{yu2023language} restricts the possible outputs of the LLM to a set of functions and parameters from a predefined library. 
Moreover, the framework in \cite{yu2023language} does not allow for constraint inclusion, which has been shown to boost performance and increase system safety \cite{sharma2022correcting}. 

Reward functions generated by LLMs have also been used to train Reinforcement Learning policies in order not to rely explicitly on the LLM for motor-control tasks \cite{ma2023eureka}. Despite these policies being better suited to deal with the accuracy-robustness tradeoff, they still do not allow for the incorporation of constraints and introduce significant delays due to their training time. Furthermore, the successful deployment of these policies in the real world is particularly jeopardized by the sim-to-real gap. \looseness=-1 

\section{Problem Statement}

\begin{figure*}[h]
\centering
\includegraphics[width=\linewidth,clip]{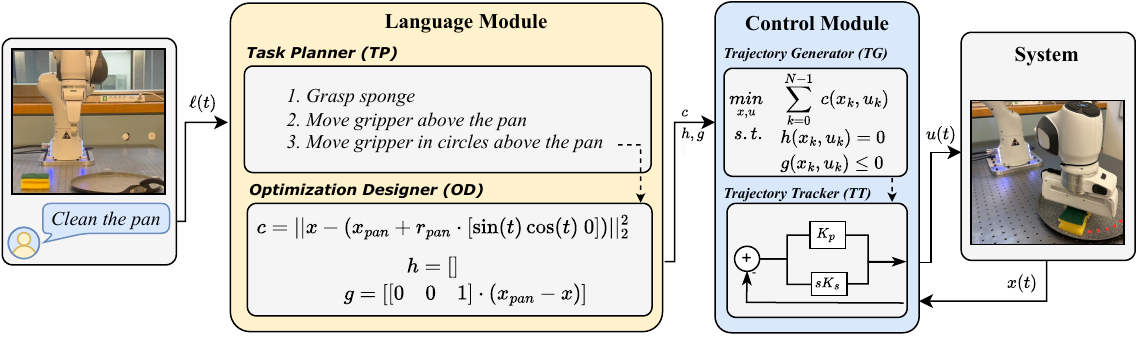}
    \caption{\textbf{Proposed Architecture:} Overall architecture for the control of a robot manipulator via natural language. The user provides a task in natural language $\ell$, which then gets translated into a series of steps (TP) and objective and constraints (OD) via two layered blocks of a language model in the language module. The objective and constraints $c,h,g$ are then used as inputs to the control module, which generates a trajectory via MPC (TG) and the low-level control commands (TT) to be applied to the system, in this case, a robotic arm.}
    \label{fig: nlc_architecture} \vspace{-4px}
\end{figure*}

Consider a discrete-time system at time $t$ with dynamics
\begin{equation}\label{eqn: dynamical_system}
    x(t+1) = f(u(t),x(t)),
\end{equation}
where $x(t)\in\reals{N_x}$ is the state of the system and $u(t)\in\reals{N_u}$ is the control input.

We introduce the \emph{natural language control} problem. 
Here, the control input $u(t)$ is a function of a natural language string $\ell(t)$ given at time $t$, as well as the current state of the system $x(t)$. 
Hence, we study the problem of designing $K$ such that 
\begin{equation}\label{eqn: natural_language_controller}
    u(t) = K(x(t),\ell(t), t),
\end{equation}
where $\ell(t) \in\Sigma^*$ for all $t$ represents the string of natural language instructions over an alphabet\footnote{An alphabet is a finite, non-empty set. 
The Kleene closure of an alphabet $\Sigma^*$ is defined as the following infinite union $\bigcup_{n=0}^\infty \Sigma^n$ where $\Sigma^n$ is the $n$-times Cartesian product, i.e., all strings of length $n$.} 
$\Sigma$ and $K:\ \Sigma^* \times \mathbb R^{N_x} \times \mathbb R \rightarrow \mathbb R^{N_u}$. 
Notice that a time-varying control policy is allowed in this setup.\looseness=-1

Since the natural language control problem formulation posed in problem \eqref{eqn: natural_language_controller} is broad, we restrict ourselves to the optimal control setting where we can restrict the set of safe trajectories via hard constraints. 
To that end, we introduce the following assumption.\looseness=-1

\textit{Natural language MPC-based control:} We assume that the natural language commands $\ell(t)$ given to the robot encode a goal that can be achieved via an optimal control problem. In particular, we assume that the goal of the controller $K$ in equation \eqref{eqn: natural_language_controller} is to design a trajectory ($x_1,\ldots,x_N; u_1,\ldots,u_{N-1})$ that optimizes an objective function (cost function), while also being subject to $n_g + n_h$ hard constraints on the state and the input. Hence, the expression for $K$ at time $t$ is implicitly encoded through the MPC subroutine
    \begin{subequations}\label{eqn:nmpc}
    \begin{align}
    \underset{\begin{subarray}{c}
      x_1,\ldots,x_N \\ u_1,\ldots,u_{N-1}\end{subarray}}{\text{min}} & \qquad {\sum_{k=0}^{N-1}}\ c(x_k, u_k;\ell(t)) \label{eqn:nmpc_cost}\\
    s.t. & \qquad x_{k+1} = f(x_k, u_k) \label{eqn:nmpc_dyanimcs} \\
    &  \qquad h(x_k, u_k;\ell(t)) = {0}
    \label{eqn:nmpc_equalities} \\
    & \qquad g(x_k, u_k;\ell(t))\leq {0}
    \label{eqn:nmpc_inequalities} \\
    & \qquad x_0 =  x(t) \label{eqn:nmpc_init}
    \end{align}
    \end{subequations} 
Note the cost function $c(x_k, u_k;\ell(t)) \in \mathbb R$, the constraints $h(x_k, u_k;\ell(t)) \in \mathbb R^{n_h}$ and $g(x_k, u_k;\ell(t)) \in \mathbb R^{n_g}$ are specified via natural language through the string $\ell(t)$.\looseness=-1


The goal of this paper is to create a system's architecture that captures the simplicity and flexibility of natural language instructions while facilitating a concrete and safe execution of the predefined task. 
To do so, we leverage pre-trained models in natural language as well as optimal control techniques and study how to best interface these systems to achieve such goals. 


\section{Language to Control}
To solve the \emph{natural language control} problem, we make use of the capabilities of a pre-trained LLM  (GPT4 \cite{openai2023gpt4}) and well-established optimal control techniques (MPC and impedance control). Our architecture consists of a planning and control module, each of them containing two blocks.
An overview of the architectural components can be seen in \cref{fig: nlc_architecture}. 
We note that this architecture is fairly general and could be applied to a variety of environments where control actions are needed to steer the behavior of a dynamic system. 
In this work, feedback in the form of robot states as well as information about the position of objects in the scene is assumed to be known and available to the Control Module at all times. However, our implementation on real robot manipulators shows that \ourmethod performs robustly even under disturbances and imperfect perception.
The scene description (i.e. available objects) is appended to each instruction sent by the user while information about the robot and its state are part of the system prompt provided to the Language Module at the beginning of the episode.

\subsection{Language Module}

The Language Module, $L:\kleene{\alphabet}\rightarrow \C{1}\times\dots\times \C{1}$,\footnote{Due to the lack of explainability of current language models, we cannot guarantee that the output of map $L$ will sit in $\C{1}\times \dots\times \C{1}$. In those cases, task execution will fail.} is designed to receive a language utterance $\lvalue{t}$ as input at each time step $t$, and generate the cost and constraints functions $c$, $h$ and $g$ as an output, i.e.,
\begin{equation}
(c,\underbrace{h_1,\dots,h_{n_h}}_h,\underbrace{g_1,\dots,g_{n_g}}_g) = L(\lvalue{t}).
\end{equation}
The goal of this module is to convert instructions given in natural language into an output that can readily be used by MPC controllers, which are frequently used for controlling dynamic systems, such as robots, in common practice. Since human instructions often lack specificity and can be ambiguous, we divide this task into two different steps. In fact, distributing one comprehensive instruction across multiple precisely prompted LLMs has been shown to significantly increase performance compared to offloading it entirely on one model prompted generally \cite{wang2023voyager}. Hence, the Language Module in the proposed architecture consists of two layered blocks: the Task Planner and the Optimization Designer.

\textbf{Task Planner (TP)}:
The Task Planner block receives the user instructions as the input and generates a plan of actions that the system is required to execute. In particular, we use a pre-trained LLM as a decoder to convert the user message into a list of specific subtasks formulated in natural language that need to be carried out in order to fulfill the original task. To do this, the language model is prompted with information about the system to be controlled (i.e., a robotic arm, etc.), the environment (i.e., the presence of different objects), and general guidelines on how to generate the sequence of subtasks (few-shot examples).\looseness=-1

\textbf{Optimization Designer (OD)}:
The Optimization Designer block consists of an LLM that receives a given subtask from TP and generates appropriate objective and constraint functions $c, h, g$. In this block, we sequentially use each of the subtasks and return an optimization formulation in mathematical terms as an output. 
To do this, we rely on the ability of pre-trained models to code optimization functions symbolically using CasADi \cite{Andersson2019} (a non-linear optimization framework) as shown in a representative example in \cref{fig: cook_steak}. This provides our system with a greater versatility as compared to prior works, where the language model is often constrained to use predefined parametrized functions \cite{yu2023language}. 
Besides symbolic functions in CasADi, the language model also has been instructed in the availability of separate functions specific for the task, i.e., in the case of robot control: $\texttt{open\_gripper()}, \texttt{close\_gripper()}$. 

In the cube stacking example, as introduced in \cref{sec:tasks}, the \emph{OD} block is able to formulate objective and constraints that ensure an optimal trajectory while avoiding collisions with other cubes. In this case, we define the state $x\in\mathbb R^4$ as the 3D position and orientation $\psi$ around the z-axis of the robot gripper, and the control inputs $u \in\mathbb R^4$ as the time derivative of the states ($u = \dot{x}$). This choice is conveyed to the \emph{OD} block through an instruction in natural language. 

As an example, the natural language utterance $\ell = $``\textit{move on top of the blue cube avoiding collisions with red and green cubes}'' gets translated into the following objective function\looseness=-1
\begin{equation}\label{eqn: OD_cost_example}
    c(\ell,x.u) = \left \|\text{\small diag(1,1,1,0)}(x - x_{\text{blue}} + [0, 0, d])\right \|_2^2
\end{equation}    
and the following collision-avoidance constraints
\begin{subequations}\label{eqn:OD_constraints_example}
\begin{align}\label{eqn: OD_constraints_example}
    & g_1(\ell ,x, u) =  d_{\text{min}} - \left \|\text{\small diag(1,1,1,0)} (x - x_{\text{red\ }})\right \|_2  \leq 0 \\
    & g_2(\ell ,x, u) = d_{\text{min}} - \left \|\text{\small diag(1,1,1,0)} (x - x_{\text{green}})\right \|_2 \leq 0 
\end{align}
\end{subequations} 

where $x_{\text{blue}}, x_{\text{red}}, x_{\text{green}}\in\mathbb R^4$ are position and orientation of the corresponding cubes, $d\in\mathbb R$ is the side length of the cubes, and $d_{min}$ is the minimum distance to avoid collisions, in this example, $d_{min} \defeq d/2$. 

This process of generating a mathematical representation given a concrete sub-task expressed in natural language is repeated until all sub-tasks are carried out. Proceeding to the next sub-task occurs if one of three conditions is satisfied: 
\begin{equation}
    {J(t) \leq \epsilon_1, \quad
    \Delta{J}(t) \leq \epsilon_2, \quad
    t - t_0 \geq t_{max} }
\end{equation}
Where $J(t)$ represents the value of the MPC objective at the current time step and $\Delta{J}(t)$ represents its difference from one step to the next. $t_0$ represents the first time step where the new MPC formulation has been applied and $\epsilon_1, \epsilon_2, t_{max}$ represents the limit thresholds that are to be tuned. 
  
\subsection{Control Module}

The control module, $C:\reals{N_x}\times  \C{1}\times\dots\times \C{1}\rightarrow \reals{N_u}$, receives the objective function and constraints $c$, $h$ and $g$ together with the current system state as $\x{t}$ as an input, and generates an optimal control signal $u$ as an output, i.e.
\begin{equation}
 \uvalue{t} = C(\x{t};c,h, g).
\end{equation}
The goal of this module is to ensure that the system (in this case the robotic arm) generates and follows an optimal trajectory while satisfying constraints. Once again, we divide this task into two different steps.
This separation has been shown in the literature to maximize the system performance by producing accurate high-level motions that can be tracked at high frequencies by a low-level controller and in our architecture is presented as two layered blocks: the Trajectory Generator and the Trajectory Tracker.

\textbf{Trajectory Generator (TG)}:
The Trajectory Generator block consists of an MPC controller that receives the objective function $c$ and constraints $h, g$ from the \emph{OD} and solves for an optimal trajectory, i.e., sequence of states and inputs $x_n$ and $u_n$ for $n=1,\dots,N$ where $N$ is some predetermined time horizon. 
The TG solves the MPC subroutine \eqref{eqn:nmpc} at every time step $t$ for the new measurement  $x(t)$ using a predefined system model \eqref{eqn:nmpc_dyanimcs}. 
Usually, a simplified model is used at the TG level to reduce latency and modeling complexity. In this way, it is still possible to generate trajectories that minimize the objective function while also respecting the constraints received from the LLM without adding modeling overhead at the planning level. 
The separation of TG and TT specifically allows us to deal with this issue by relying on a much simpler approximation of the robot manipulator system, i.e.
\begin{equation}\label{eqn:robot_dynamics}
x(t+1) = A\x{t}+B\uvalue{t},
\end{equation}
where the diagonal matrices $A = I_4$ and $B = I_4 \cdot \Delta t $ represent the kinematics matrices of a point mass in 3D position and rotation around the z-axis of the end-effector with $I$ and $\Delta t$ representing the identity matrix and discretization time respectively. 
Additions to the MPC, such as regularization terms and pre-defined constraints, can be appended in an additive manner to the objective function and constraints received from the OD. We introduce safety constraints on the end-effector state $x \in \mathbb R^4$ to be within  a feasible set (i.e. above the table surface) as
\begin{equation}
\label{eqn:OD_apriori_constraints_example}
    g_1(x,u) = \begin{bmatrix} 0 & 0 & -
    1 & 0 \end{bmatrix}  x \leq 0.
\end{equation}
We also introduce safety constraints to bound the gripper velocity $u \in \mathbb R$ in a similar fashion to \eqref{eqn:OD_apriori_constraints_example}. Hence, the TG for all tasks described in \ref{sec:tasks} solves subroutine \eqref{eqn:nmpc}, where the dynamics \eqref{eqn:nmpc_dyanimcs} are replaced by expression \eqref{eqn:robot_dynamics}, and constraint \eqref{eqn:OD_apriori_constraints_example} is included as part of the formulation \eqref{eqn:nmpc_inequalities}.

The choice of MPC for the TG block is motivated by the simplicity of its formulation, together with the computation efficiency and the ability to provide guarantees on the generated trajectory. While similar approaches make use of the generated objective (reward) function to train learning-based controllers \cite{suhr2022continual, ma2023eureka}, they require the controller to be trained for each task independently, which severely limits their generalizability and makes it challenging to use them in a real-world setting. 
The ability of LLMs to account for human feedback alongside the real-time regime of MPC, on the other hand, enables interactive collaboration between human users and NARRATE, which can further improve task performance as shown in  \cref{sec: results}.


\textbf{Trajectory Tracker (TT)}:
The Trajectory Tracker block is the lowest-level block of the architecture stack and aims to follow the trajectory generated by the TG as optimally as possible. Its purpose is to map the trajectories obtained from TT into specific low-level actions for the given hardware, i.e., torque inputs to be applied to the motors of a robot, etc. 
Once the TT receives the optimal trajectory $\left[x^*_1,\dots,x^*_N, u^*_1,\dots,u^*_{N-1}\right]$ from the TG, it computes the desired acceleration of the gripper $w$ via a cartesian impedance controller:
 \begin{subequations}\label{eqn:PD_controller}
\begin{align} 
\dot{w}_k^* &= k_p (x_k^*-x(k)) + k_d (u_k^*-\dot{x}(k)), \\
\tau_k &= J_e^T (\Lambda \Dot{w}_k^* + \mu + r)
\end{align}
 \end{subequations}
%
    
where $k_p$ and $k_d$ represent the gains of the PD controller, respectively, and $x(k), \dot{x}(k), x_k^*, u_k^*$ represent the measured and desired gripper position and velocity at time $k$, respectively. $J_e$, $\Lambda, \mu, r$ are obtained from the dynamics model as shown in \cite{Baruh1998-vf}.


\begin{table*}
	\setlength{\tabcolsep}{3pt}
	\centering
	\begin{tabular}{l|cc|cc|cc|cc|cc|cc}
		\toprule
		\multirow{2}{*}{Method} & \multicolumn{2}{c|}{Stack} &  \multicolumn{2}{c|}{L-Shape}   &  \multicolumn{2}{c|}{Pyramid} &  \multicolumn{2}{c|}{Clean Plate} &  \multicolumn{2}{c|}{Move Wet} &  \multicolumn{2}{c}{Cook Steak} \\ 
		                                  & SR$_\%$$\uparrow$ & Co/Pl/Cd$_\%$$\downarrow$     & SR$_\%$$\uparrow$ & Co/Pl/Cd$_\%$$\downarrow$ & SR$_\%$$\uparrow$ & Co/Pl/Cd$_\%$$\downarrow$ & SR$_\%$$\uparrow$ & Co/Pl/Cd$_\%$$\downarrow$ & SR$_\%$$\uparrow$ & Co/Pl/Cd$_\%$$\downarrow$ & SR$_\%$$\uparrow$ & Co/Pl/Cd$_\%$$\downarrow$ \\ \toprule
		CaP \cite{liang2023code}   & \textbf{98}   & \ 0 / 2 /\ \ 0 & 10            & 60 / 16 / 14          & 16            & \ 2 / 76 / \ 6   & 22            & 10 / 48 / 20          & 4             & 36 / 40 / 20          & 0             & 20 / 48 / 32          \\ 
		VoxPoser \cite{huang2023voxposer} & 26            & 48 / 0 / 26    & 0             & 76 / \ 0 / 24         & 16            & 38 / 22 / 24     & 48            & 6 / 22 / 24           & 6             & \ 10 / 24 / 60           & 0             & \ 0 / 74 / 26           \\ 
		NARRATE$^\dagger$            & 50            & 48 / 2 /\ \ 0  & 24            & 66 / 10 / 0           & 32            & 58 / 10 / \ 0    & \textbf{64}   & 10 / 26 / 0           & 64            & 18 / 18 / 0           & 0             & 50 / 50 / \ 0           \\
		\ourmethod                        & 92            & \ 2 / 6 /\ \ 0 & \textbf{58}   & 12 / 30 / \ 0           & \textbf{76}   & \ 2 / 22 / \ 0   & 62            & \ 4 / 34 / 0            & \textbf{68}   & 16 / 16 / 0           & \textbf{30}   & 30 / 40 / \ 0           \\ \bottomrule
	\end{tabular}
	\caption{\textbf{Simulation Results:} Comparison of \ourmethod against \emph{CaP} and \emph{VoxPoser}  on the six simulation tasks as illustrated in \Cref{fig: all_tasks}. We report the success rates for each method alongside the failure reason categorized into Collision (\emph{Co}), Planning (\emph{Pl}) and Code Execution (\emph{Cd}) errors. $\dagger$ represents the version without constraints.\vspace{-12px}}
	\label{tab:sucess_rates}
\end{table*}
\section{Experimental Setup}
We extensively evaluate our method in custom simulation environments using PandaGym \cite{gallouedec2021pandagym} and on two real-world platforms consisting of the Franka Emika Panda and an in-house built manipulator (DynaArm). 

\begin{figure}[h!]
    \centering

    \begin{subfigure}[b]{0.32\linewidth}
    \centering
            \begin{tikzpicture}
            \node at (0, 0) {\includegraphics[height=3.8cm]{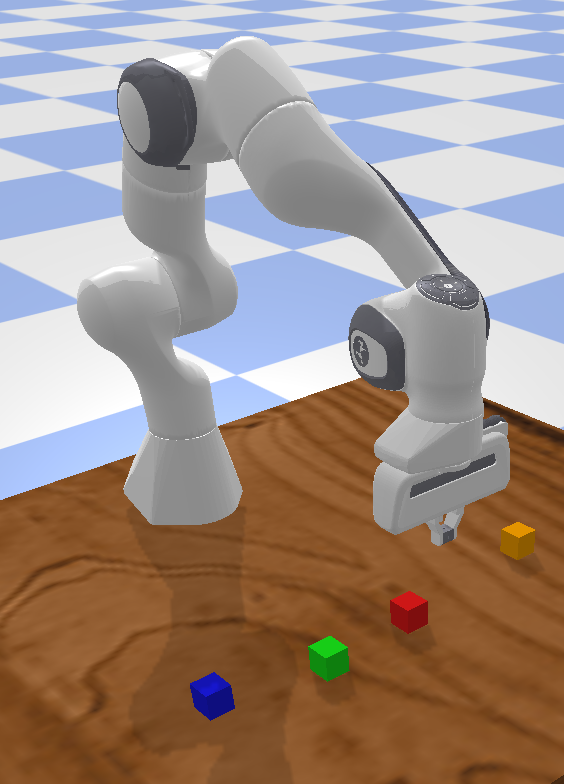}};
            \end{tikzpicture}
            \caption{Simulation \cite{gallouedec2021pandagym}}
    \label{fig:experiment_setup_sim}
            
    \end{subfigure}
    \begin{subfigure}[b]{0.66\linewidth}
    
    \centering
            \begin{tikzpicture}
            \node at (0, 0) {\includegraphics[height=3.8cm]{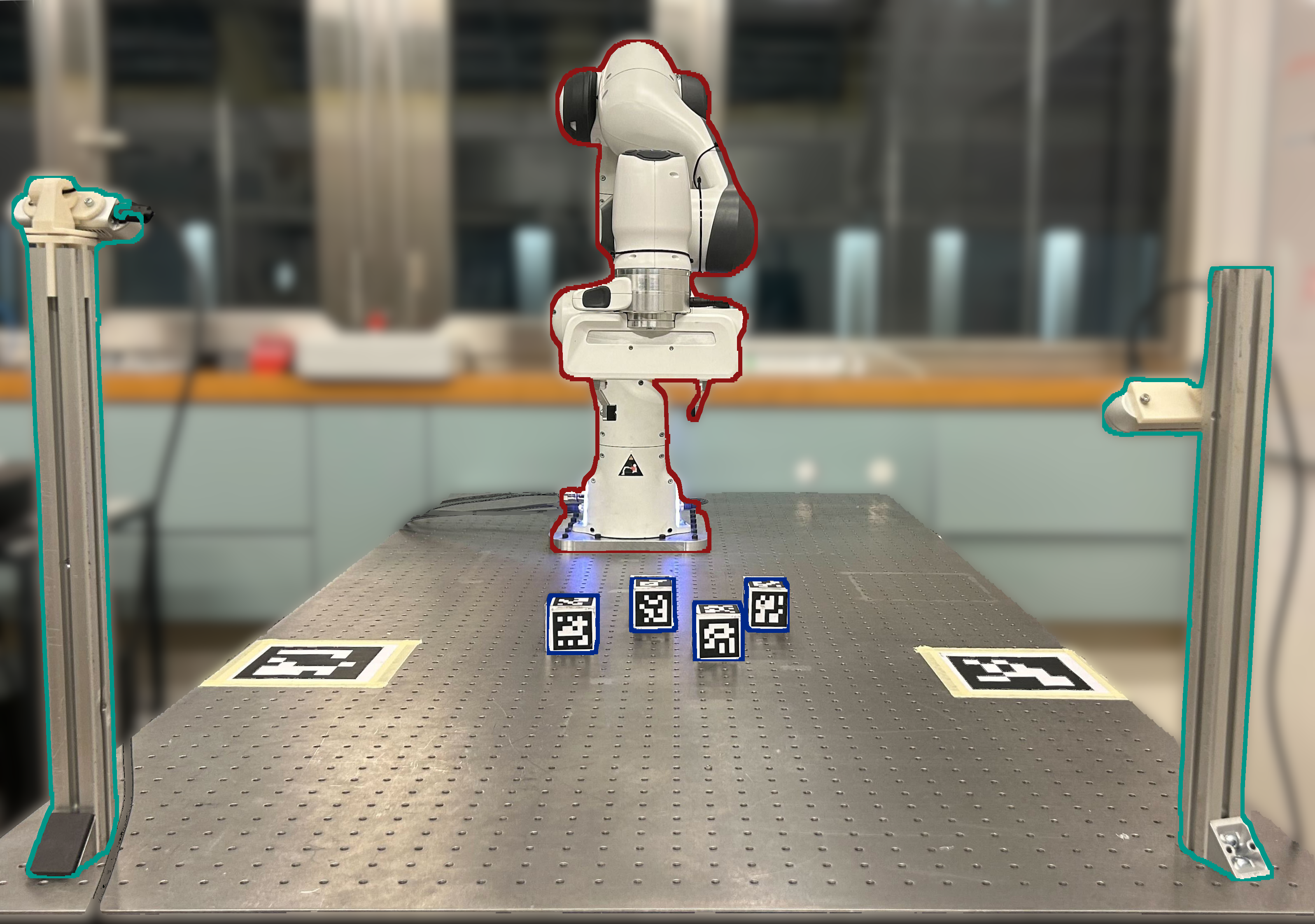}};
            \node[anchor=east] (cam) at (1.2, -1.5)  {\footnotesize RGB-D Cameras};
            \draw[blue, thick] (cam) -- (-2.5, 0.9);
            \draw[blue, thick] (cam) -- (1.7, 0.1);
            \end{tikzpicture}
            \caption{Hardware}
    \label{fig:experiment_setup_real}
            
    \end{subfigure}
    \caption{\textbf{Evaluation setup} We use PandaGym \cite{gallouedec2021pandagym} to build our custom simulation environments. For the quantitative, real-world evaluation, we use a Franka Emika Panda with two RGB-D cameras (D435i, D455) and ArUco markers to extract object poses. 
    \vspace{-12px}}
 \end{figure}
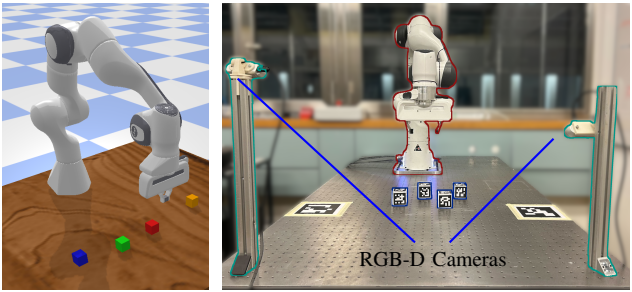

\subsection{Tasks}
\label{sec:tasks}
We evaluate our method on a total of six different tasks with varying difficulties. The first three tasks (\emph{Stack}, \emph{Pyramid}, \emph{L}) involve the manipulation of cubes by an individual robotic arm, aiming to assess the safety and high-level reasoning capabilities of the method. 
The fourth task (\emph{Clean}) includes kitchen utensils and requires the robot to solve a contact-rich task by applying time-varying control to clean a pan with a sponge. 
Finally, the last two tasks (\emph{Move Sponge}, \emph{Cook Steak}) require the system to coordinate two robotic arms collaboratively, imposing additional difficulties on the method. The single-robot tasks are tested both in simulation as well as on the real world. An overview of the different tasks is depicted in \cref{fig: all_tasks}. 

In the following, we provide a brief overview of the tasks and refer the reader to our project website\footnote{\url{https://narrate-mpc.github.io}} for implementation details, exact prompts and benchmark environments for each task. 

\textbf{Stack, Pyramid, L:} In these tasks, the robot is asked to re-arrange four cubes according to a specific pattern. All cubes are initially placed on a flat surface, and the state of the cubes and robots are known throughout the full experiment. The task is solved successfully if all 4 cubes are stacked on top of each other, form a pyramid with 2 predefined cubes at the base and one at the top or are correctly re-arranged to form an L shape flat on the table. Illustrative examples of the queries used for these tasks are: ``\emph{stack all cubes on top of the blue one}", ``\emph{build a pyramid with red and blue cube at the base}" and ``\emph{write the letter L flat on the table}".
 
\begin{figure}[t!]
\centering
\lstdefinelanguage{ldl} {language=Octave,
otherkeywords={User:,OD:},
keywordstyle={\color{black}}}
\begin{lstlisting}[language=ldl]
TP: move both grippers above the `stove` and keep them at a constant distance between each other.
OD:{"objective": "ca.norm_2(x_left[:3]-(stove.x[:3]+[0, 0, 0.1]))**2 + ca.norm_2(x_rig[:3]ht-(stove[:3]+[0, 0, 0.1]))**2",
"equality_constraints": ["ca.norm_2(x0_left[:3] - x0_right[:3]) - ca.norm_2(x_left[:3] - x_right[:3])"],
"inequality_constraints": []}
\end{lstlisting}
    \caption{\textbf{TP-OD responses for the  \emph{Cook Steak} task}: the robots have to collaborate in order to move the frying pan above the stove using its two handles. This OD response requires the robot grippers to remain at a constant distance to prevent the pan from falling. }
    \label{fig: cook_steak} \vspace{-12px}
\end{figure}
\textbf{Clean:}
A sponge and a pan are located on a table, and the robot is instructed to clean the pan. The task is considered to be solved successfully if the robot performs a continuous motion (i.e. circles, back and forth,...) and the sponge is in contact with the plate. The query used for this task is ``\emph{clean the pan with the sponge}".

\textbf{Move Wet:} A wet sponge and a pan are located on a table that also features a sink. The task is formulated as a collaborative task, where two robots must coordinate and move the wet sponge to the sink without dropping water. The task is solved successfully if the sponge is moved to the sink while the pan is held beneath it throughout the trajectory to avoid water from dropping. This task can be solved by one robot carrying the pan while the other carrying the sponge above the pan. It is also accepted if the sponge is directly dropped into the pan first and then moved together to the sink. The query used for this task is ``\emph{move the sponge to the sink with the left robot but since it's wet make sure not to drop water using the pan}".

\textbf{Cook Steak:} A frying pan with two handles and a steak are placed on the table. Again, two robots are present, and the task needs to be solved collaboratively. The task is considered successful if the frying pan is moved on top of the burner plate and, subsequently, the steak is placed into the pan. The query used for this task is ``\emph{cook the steak}".

\subsection{Simulation Setup}
\label{sec:sim_setup} 
For each task, both the TP and OD are provided with general system prompts that remain unchanged across the single-robot tasks and across the collaborative tasks, showing the versatility of \ourmethod to generalize in different settings. Moreover, a user message that specifies the actual task is provided in the form of an utterance alongside a list of objects the robot can interact with. The OD uses symbolic variables to represent the robot states and the objects in the scene. These are converted into their corresponding values when the MPC is initialized at each times step.

\subsection{Hardware setup}
 
We evaluate our approach with real-world experiments using a Franka Emika Panda Robotic arm and two external realsense cameras, depicted in \cref{fig:experiment_setup_real}. We evaluate our method on a subset of the previously introduced task, namely \emph{Stack}, \emph{L}, \emph{Pyramid}, and \emph{Clean Plate} and rely on two external RGB-D cameras and AruCo markers to extract the poses of the cubes. 
Additionally, we demonstrate the embodiment transfer on a custom, in-house built manipulator (Dynaarm) and that our modular structure allows us to easily switch out the Trajectory Tracker to account for different robot dynamics. For franka, we use \emph{MoveIt Servoing} \cite{moveit} as the low-level tracking module, whereas the dynaarm uses joint-level MPC using OCS2 \cite{OCS2}.

\section{Results}
\label{sec: results}
\subsection{Simulation Results}
For each of the tasks introduced in \cref{sec:tasks}, we evaluate our method and report the success rate as well as the failure reason categorized into \emph{Collision}, \emph{Planning} and \emph{Code Execution} errors. A \emph{Collision} failure occurs if the run was not successful due to a collision, i.e. the gripper knocks down the stack of cubes when trying to place the next cube on top of it. A \emph{Planning} failure occurs if the task instruction is not satisfied at the end of the run, i.e. the steak is placed into the pan but the pan is not moved on top of the burner plate for \emph{Cook Steak}. Finally, a \emph{Code Execution} failure occurs if the LLM response raises a code exception when being evaluated. It is important to note that failures are not exclusive, meaning that a failure related to \emph{Code Execution} does not imply that the run would not have had \emph{Planning} or \emph{Collision} errors. 

We evaluate our architecture over a total of 50 runs for each task and compare against current state-of-the-art methods such as \emph{Code-as-Policies} (\emph{CaP}) \cite{liang2023code} and \emph{VoxPoser} \cite{huang2023voxposer}. We implement conservative pick and place and motion primitives for \emph{CaP}, while for \emph{VoxPoser}, we programmatically derive the objects' point cloud given their known geometry and pose. The final comparisons are shown in \cref{tab:sucess_rates} and we refer the reader to our project page for additional demos. We also evaluate NARRATE without the incorporation of constraints, where the OD only generates the objective function of the MPC formulation. This version falls short both in success rate and the number of collisions, highlighting the importance of incorporating constraints into the controller formulation. This is particularly evident for the \emph{Cook Steak} task, where constraints are required for the robots to collaboratively move the pan without dropping it. The OD formulation for this specific sub-task, as seen in Figure \ref{fig: cook_steak}, enforces that the distance between the grippers remains constant. This shows how constraints are not only needed for collision avoidance but are critical for designing successful control behaviors. We highlight that our method achieves the highest success rates in five out of six tasks while showing that the adoption of constraints not only significantly reduces the number of collisions but also increases performance. 

\subsection{Efficiency Evaluations}

\begin{figure*}[t!]
    \centering
   
  \renewcommand{\arraystretch}{2.0}

    \resizebox{0.9\textwidth}{!}{
            \begin{tabular}{ccccc} 
                    \includegraphics[width=.90\linewidth]{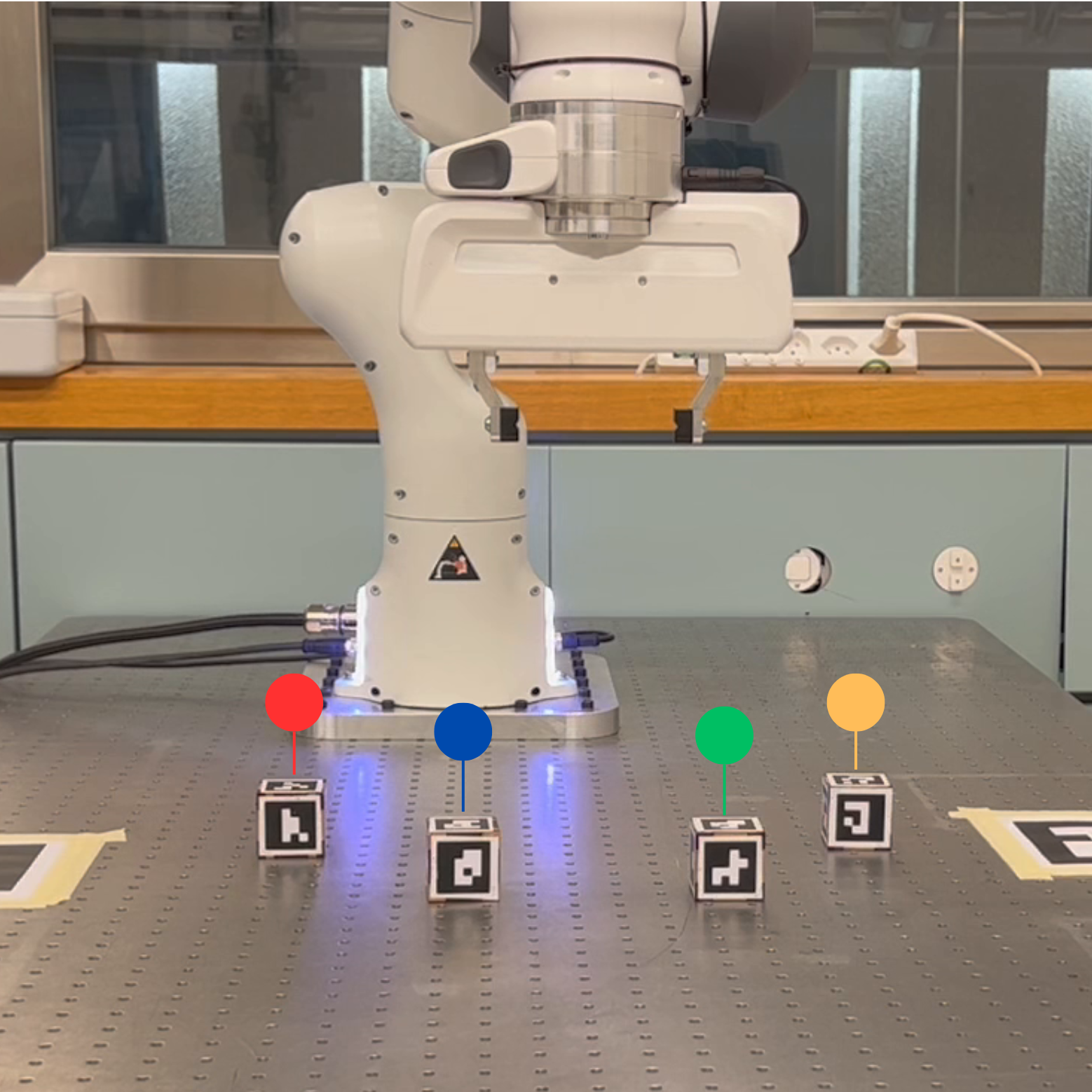}
                    &\includegraphics[width=.90\linewidth]{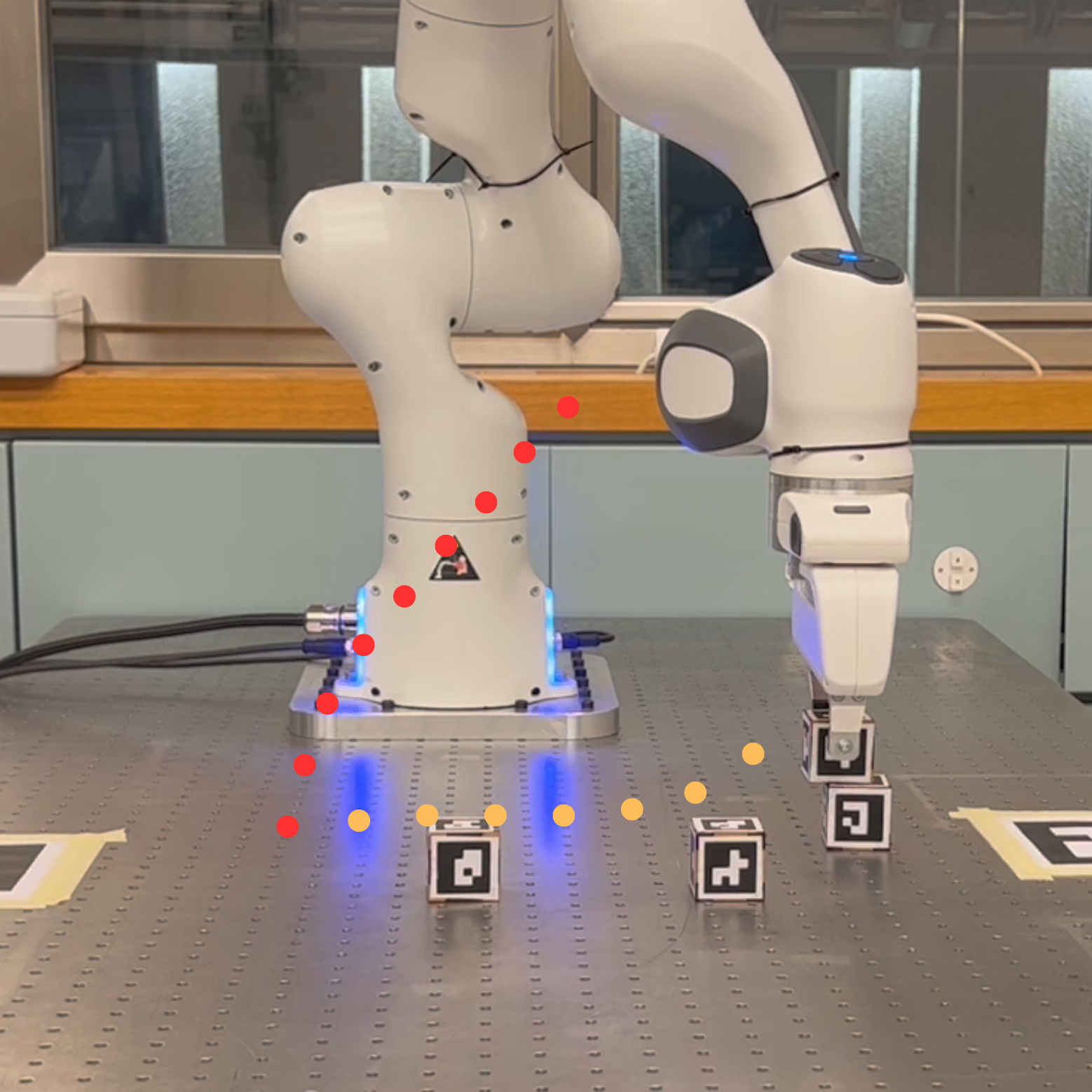}
                    &\includegraphics[width=.90\linewidth]{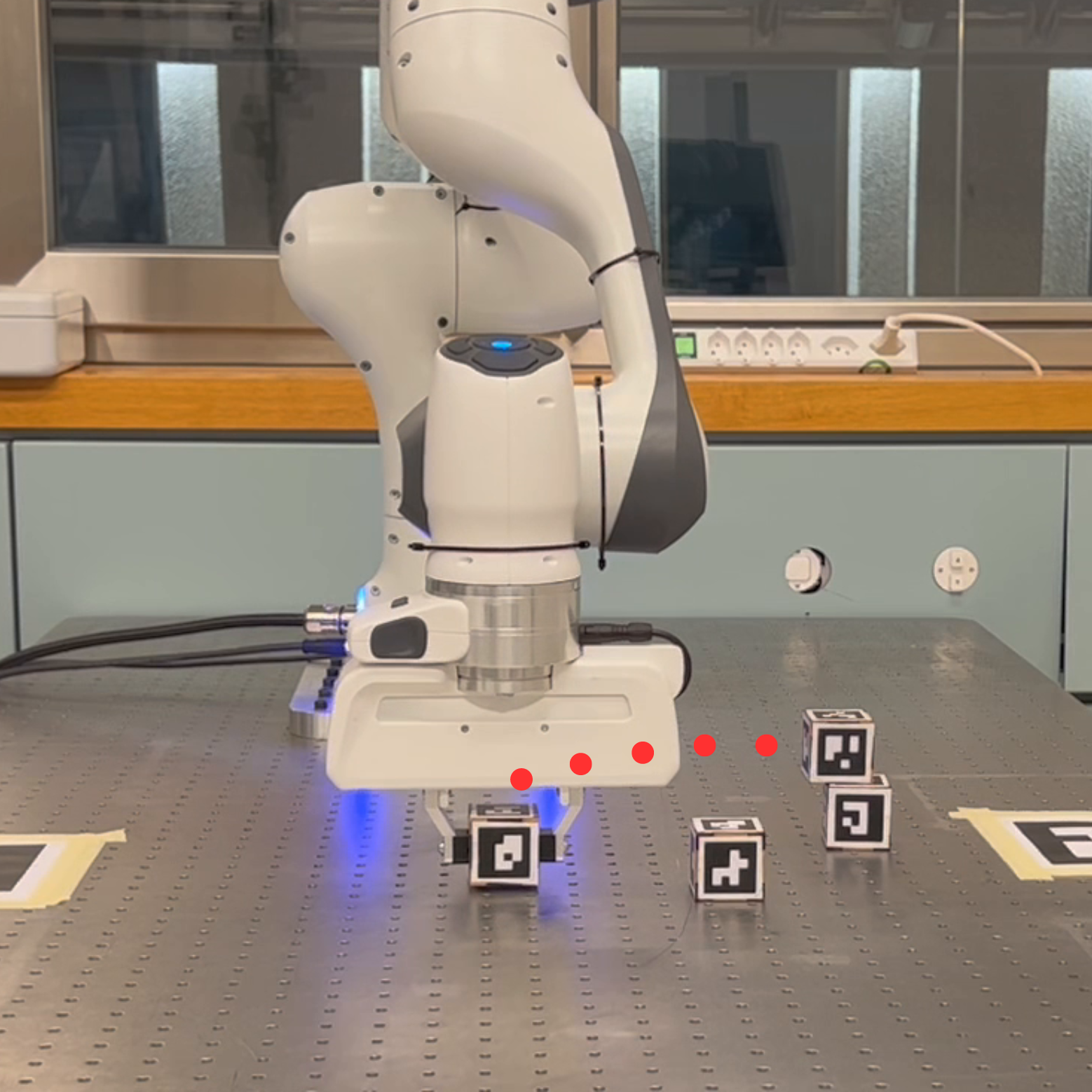}
                    &\includegraphics[width=.90\linewidth]{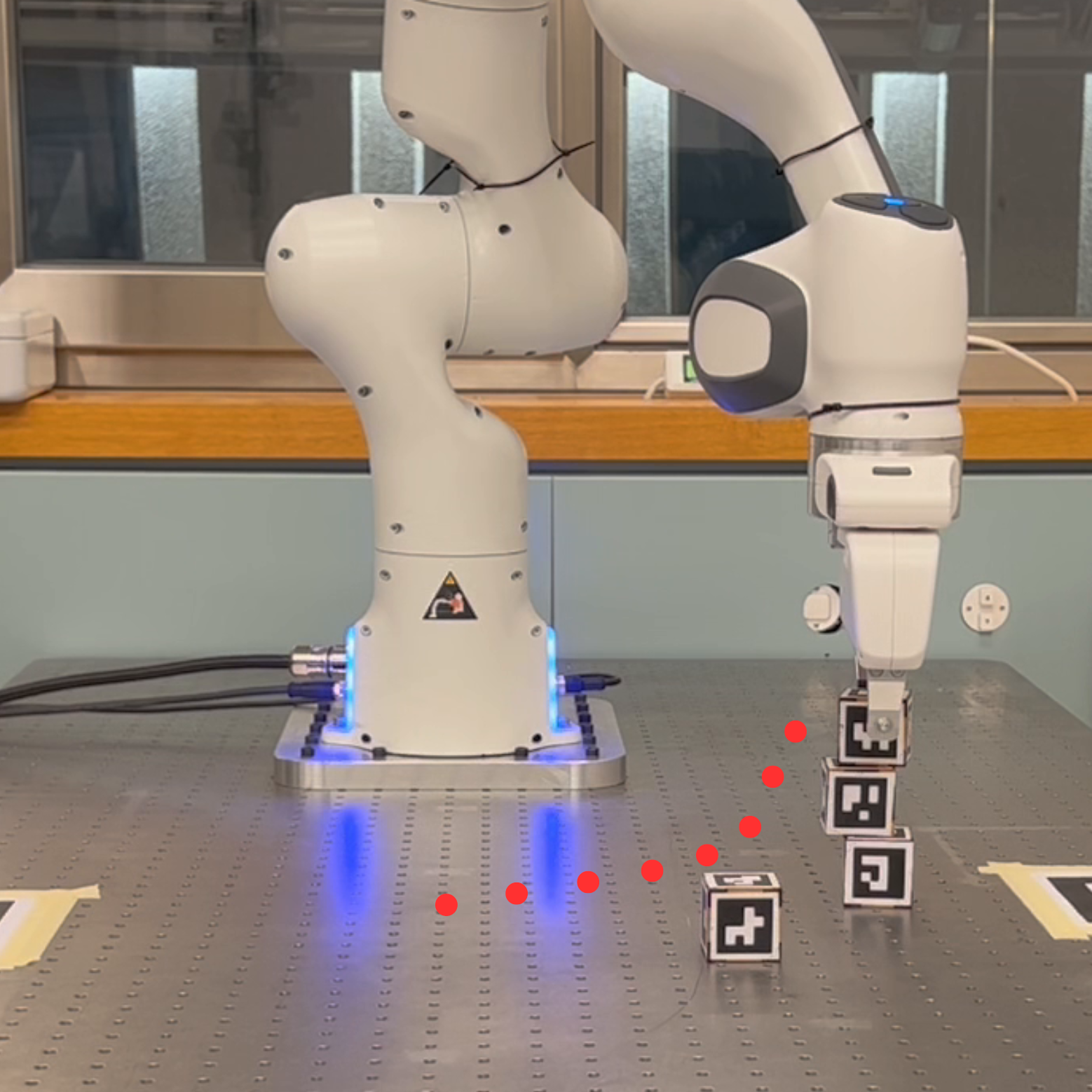}
                    &\includegraphics[width=.90\linewidth]{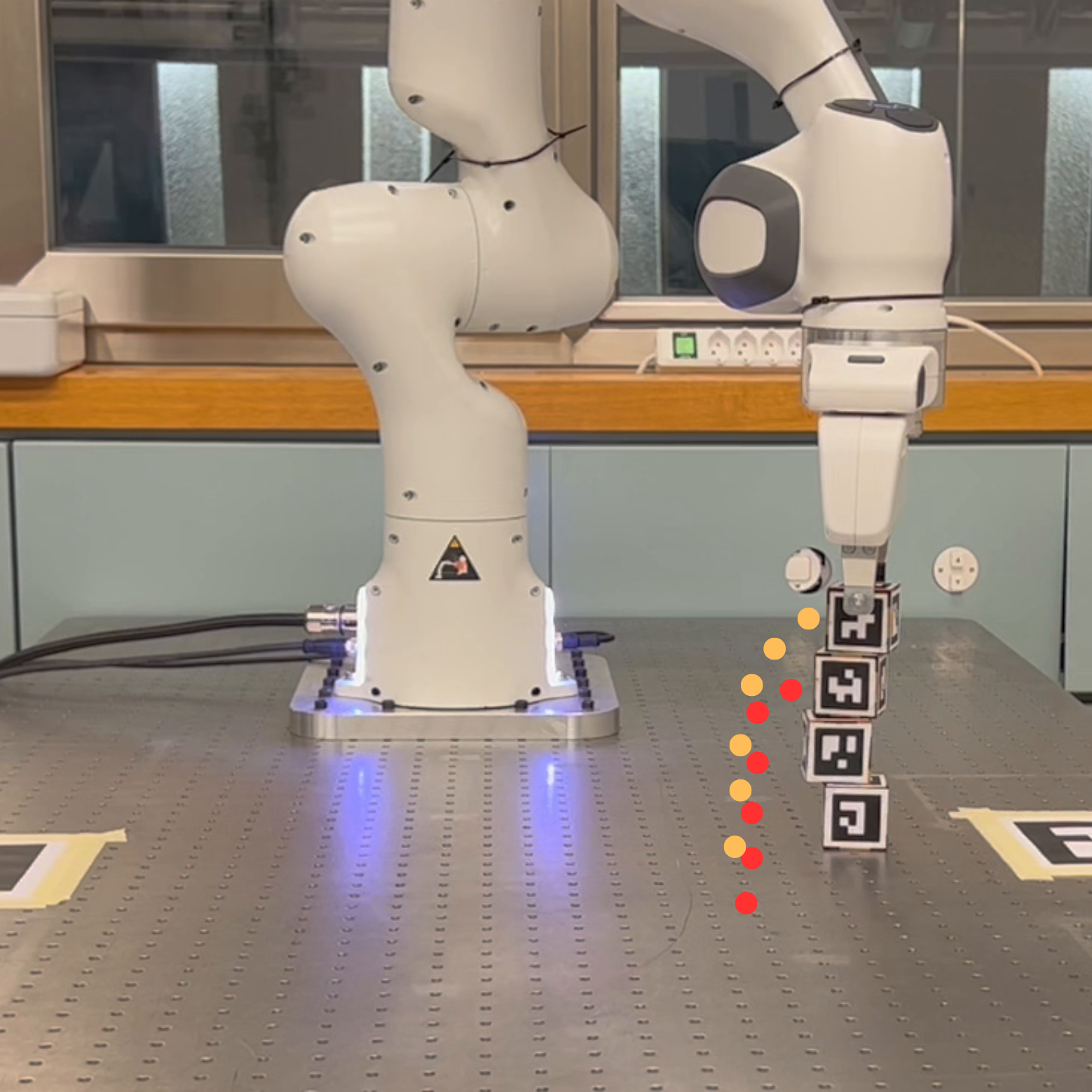}\\ \vspace{2mm}
                    \includegraphics[width=.90\linewidth]{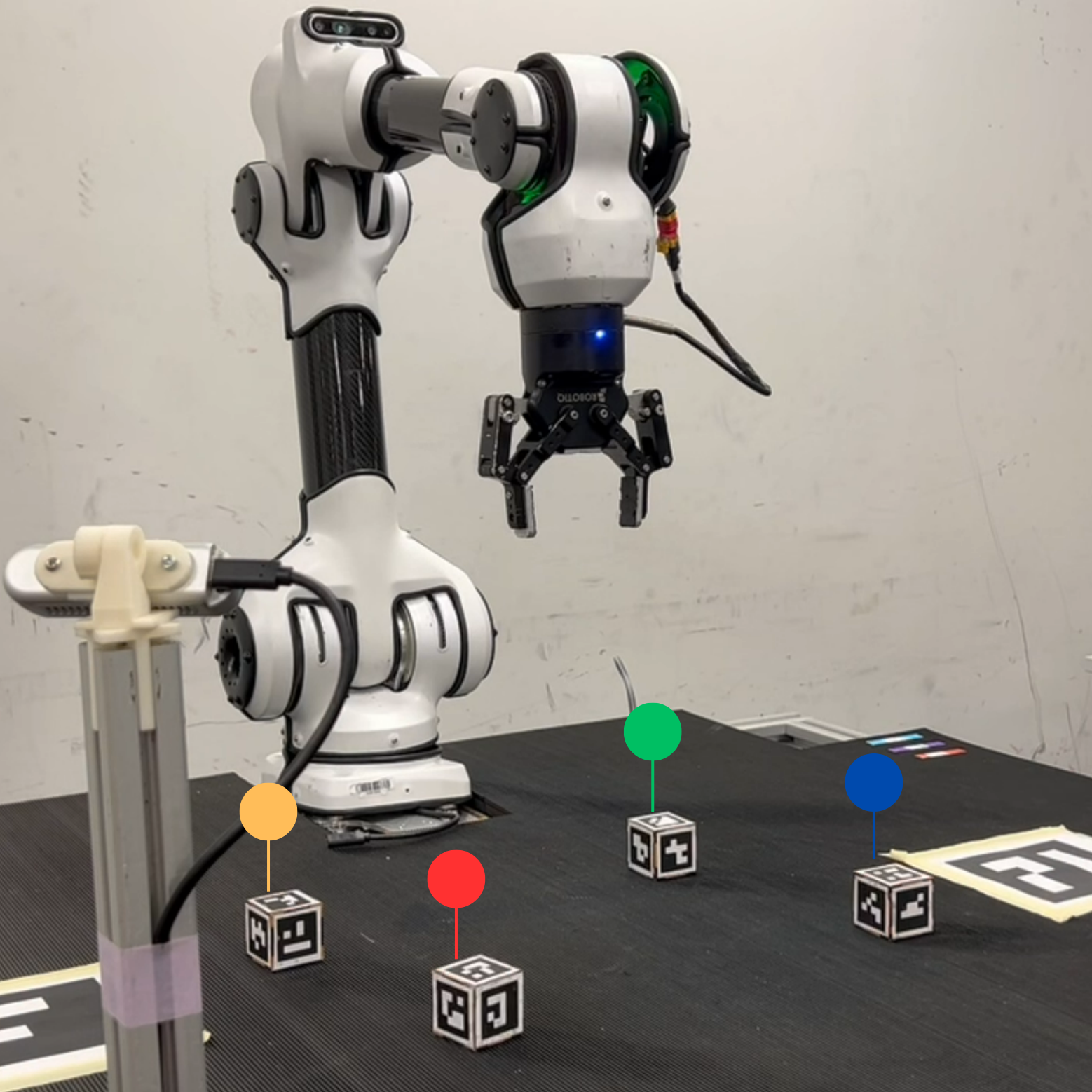}
                    &\includegraphics[width=.90\linewidth]{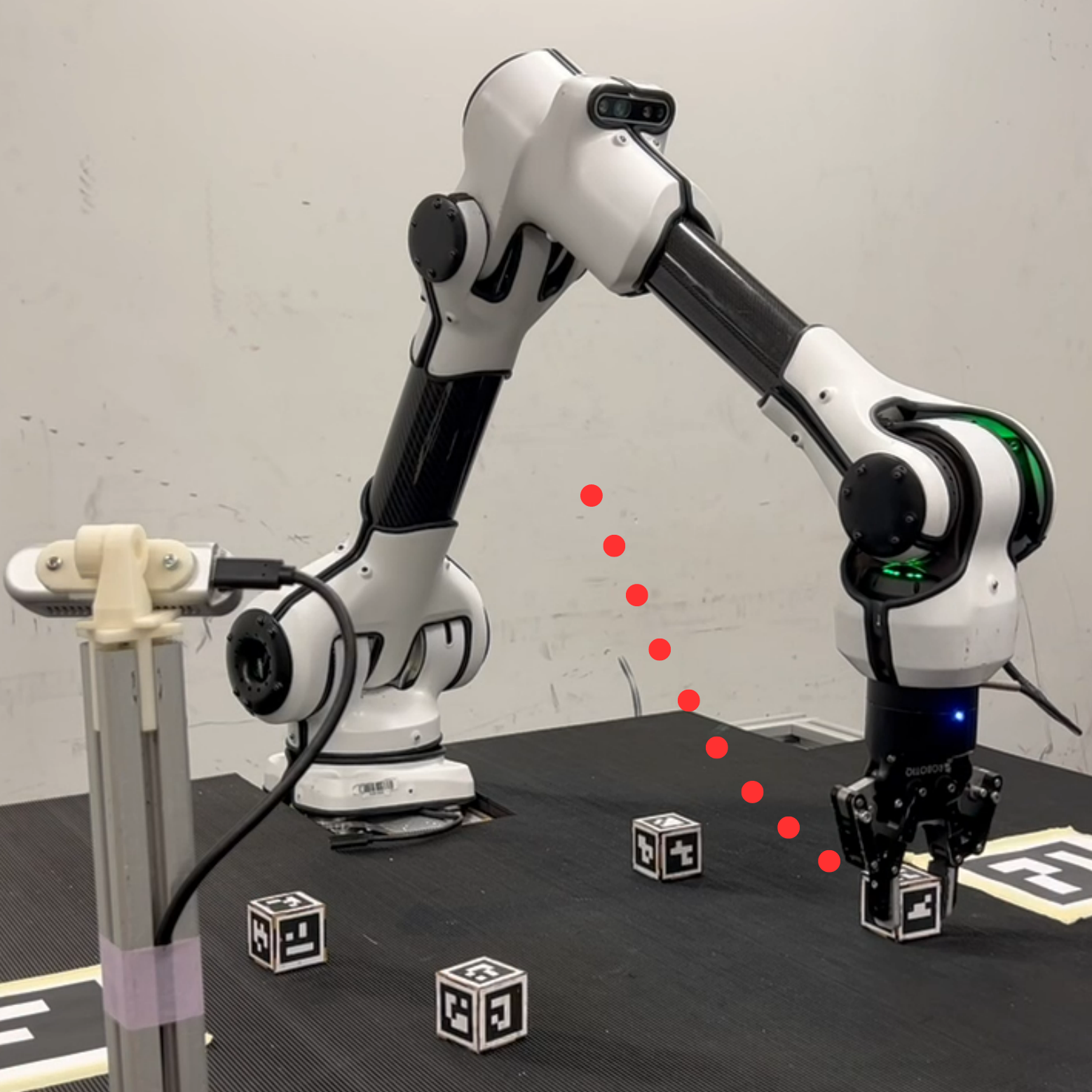}
                    &\includegraphics[width=.90\linewidth]{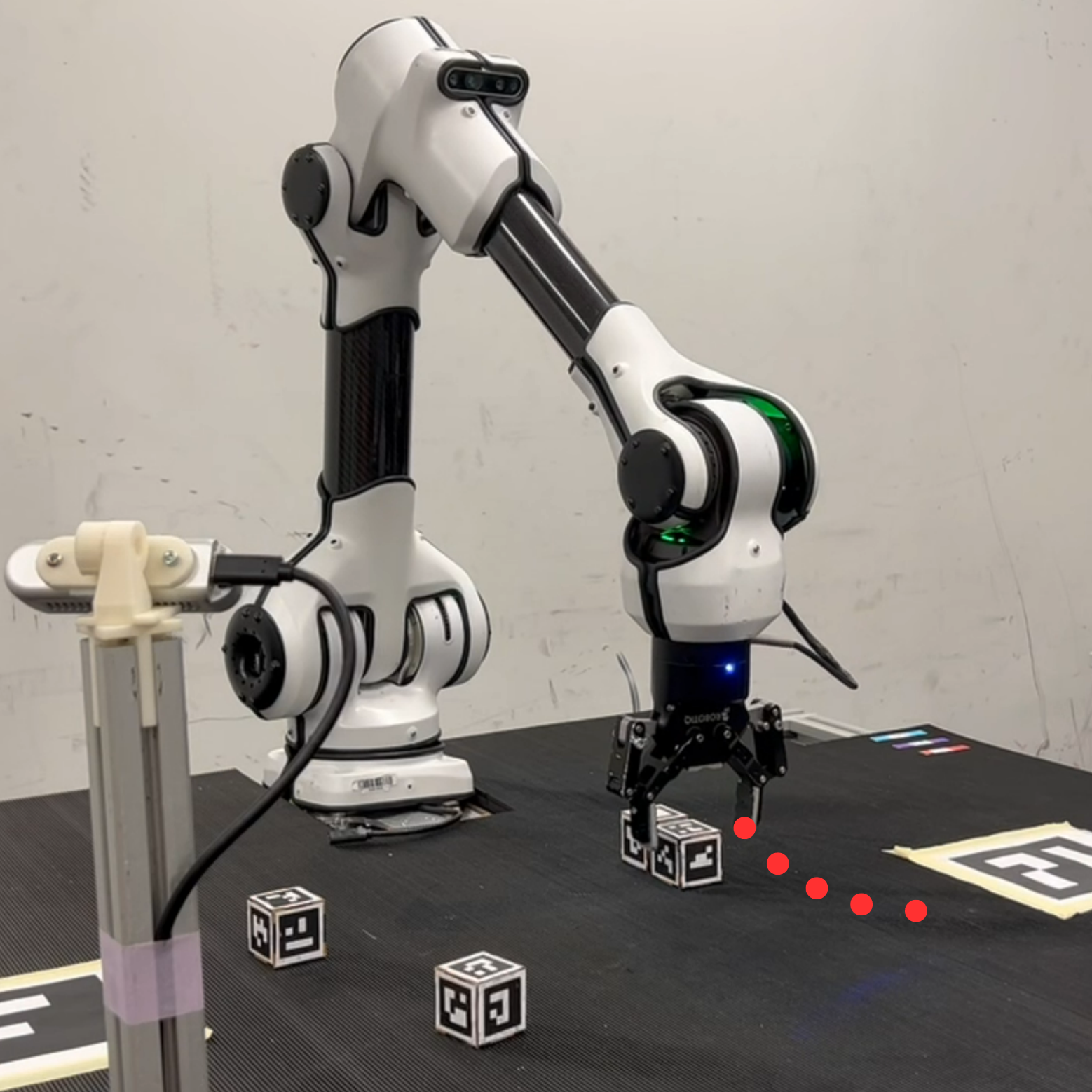}
                    &\includegraphics[width=.90\linewidth]{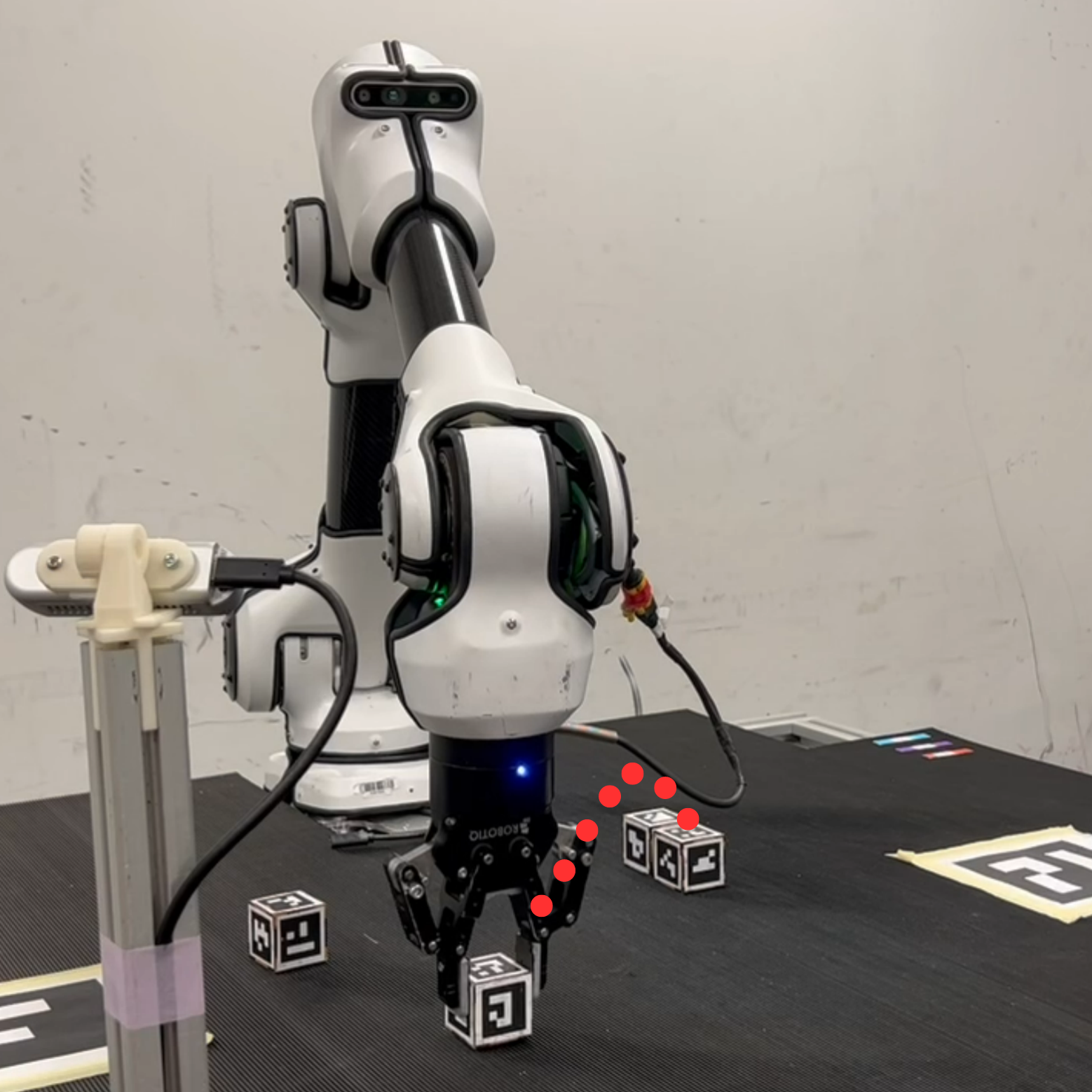}
                    &\includegraphics[width=.90\linewidth]{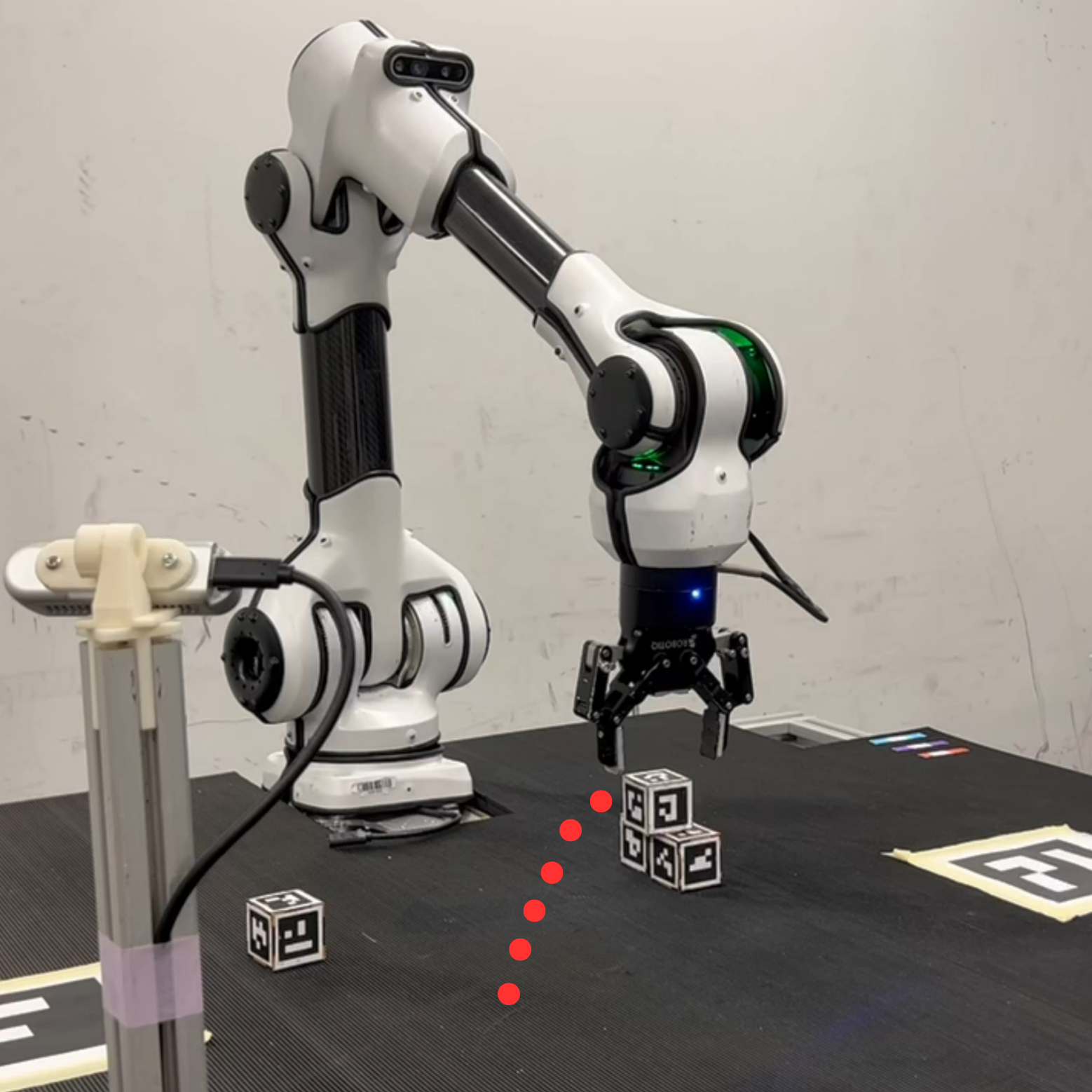}\\
            \end{tabular}}
            
    \caption{\textbf{Real World Examples:} Visualization of two real world evaluations. The respective queries for the tasks were given as (a) ``\emph{make a stack of cubes on top of the yellow cube}'' and (b) ``\emph{build a pyramid with the green and blue cubes at the base and red cube at the top. keep the green cube at its original position}''. The Aruco markers encode the color of each cube.\vspace{-12px}}
    \label{fig:real_world_rollouts}
\end{figure*}

We further evaluate the efficiency of our method by comparing the average task execution time and distance covered by the end-effector during \emph{successful} runs, as found in \cref{tab:performance_eval}. The methods that include collision avoidance are generally more efficient in terms of distance covered by the end-effector. In fact, both \emph{CaP} and NARRATE without constraints need to be prompted such that the robot first has to go on top of an object instead of directly going to its location in order to grasp it. While \emph{VoxPoser} is relatively efficient in terms of distance covered by the end-effector, thanks to the use of collision maps, its trajectory generator (a sampling-based MPC) requires significant computational efforts leading to high execution times compared to the other methods. NARRATE achieves the best results both in terms of time and distance efficiency thanks to its simple formulation and the ability to use constraints for collision avoidance. \looseness=-1

\begin{table}[h!]
\setlength{\tabcolsep}{2pt}

    \centering
    \begin{tabular}{l|cc|cc|cc|cc}
    \toprule
    \multirow{2}{*}{Method} & \multicolumn{2}{c|}{(Cubes + Clean)} &  \multicolumn{2}{c|}{Wet} &  \multicolumn{2}{c}{Steak} \\ 
    & Time$_{[\text{s}]}$$\downarrow$& Dist$_{[\text{m}]}$$\downarrow$ & Time$_{[\text{s}]}$$\downarrow$& Dist$_{[\text{m}]}$$\downarrow$ & Time$_{[\text{s}]}$$\downarrow$& Dist$_{[\text{m}]}$$\downarrow$   \\ \toprule
    CaP \cite{liang2023code} & \ 35\tiny$\pm$3  & 2.2\tiny$\pm$.1 & 188\tiny$\pm$24 & 2.6\tiny$\pm$0.5 & -- & -- \\ 
    VoxPoser \cite{huang2023voxposer} & 409\tiny$\pm$89  & 2.1\tiny$\pm$.2 & 203\tiny$\pm$17 & 1.6\tiny$\pm$0.1 & -- & --\\ 
    NARRATE$^\dagger$ & \ 38\tiny$\pm$6  &  2.5\tiny$\pm$0.2 & \ 91\tiny$\pm$11 & 1.3\tiny$\pm$0.2 & -- & --\\ 
    \ourmethod & \ \textbf{23}\tiny$\pm$\textbf{2} &  \textbf{1.7}\tiny$\pm$\textbf{0.2}&\  \textbf{65}\tiny$\pm$\textbf{7} & \textbf{1.2}\tiny$\pm$\textbf{0.1} & 91\tiny$\pm$13 & 2.0\tiny$\pm$0.2 \\ 
    \bottomrule
    \end{tabular}
    \caption{\textbf{Efficiency Evaluation:} Comparison of the efficiency of \ourmethod against \emph{CaP} and \emph{VoxPoser} as defined by the average task execution time and end-effector covered distance for \emph{succesful} runs. We average the efficiencies for the single-robot tasks and keep the others separated. 
    $\dagger$ represents the version without constraints. \vspace{-15px} }
    \label{tab:performance_eval}
\end{table}

\subsection{Continual Learning and Feedback}
\begin{figure}[b!]
\centering
\begin{subfigure}[b]{0.9\linewidth}
    \centering
    \includegraphics[width=\linewidth]{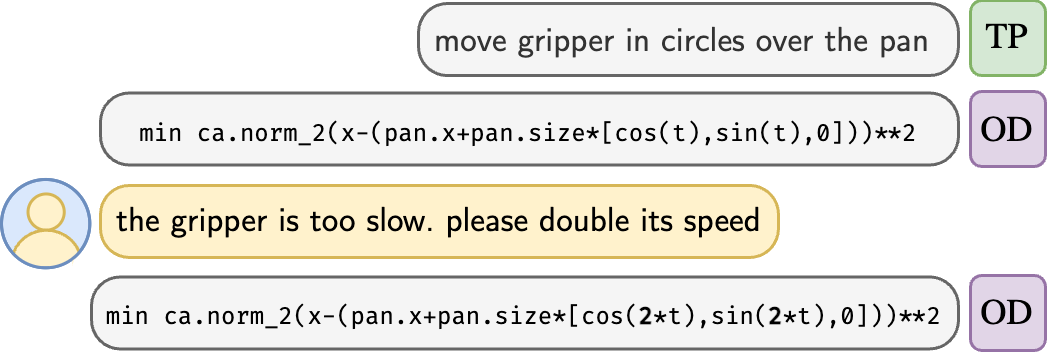}
    \caption{{Human-LLM interactive behavior}}
    \label{fig: OD_feedback}
\end{subfigure}

\vspace{5pt} 

\begin{subfigure}[b]{0.95\linewidth}
    \centering
    \includegraphics[width=\linewidth]{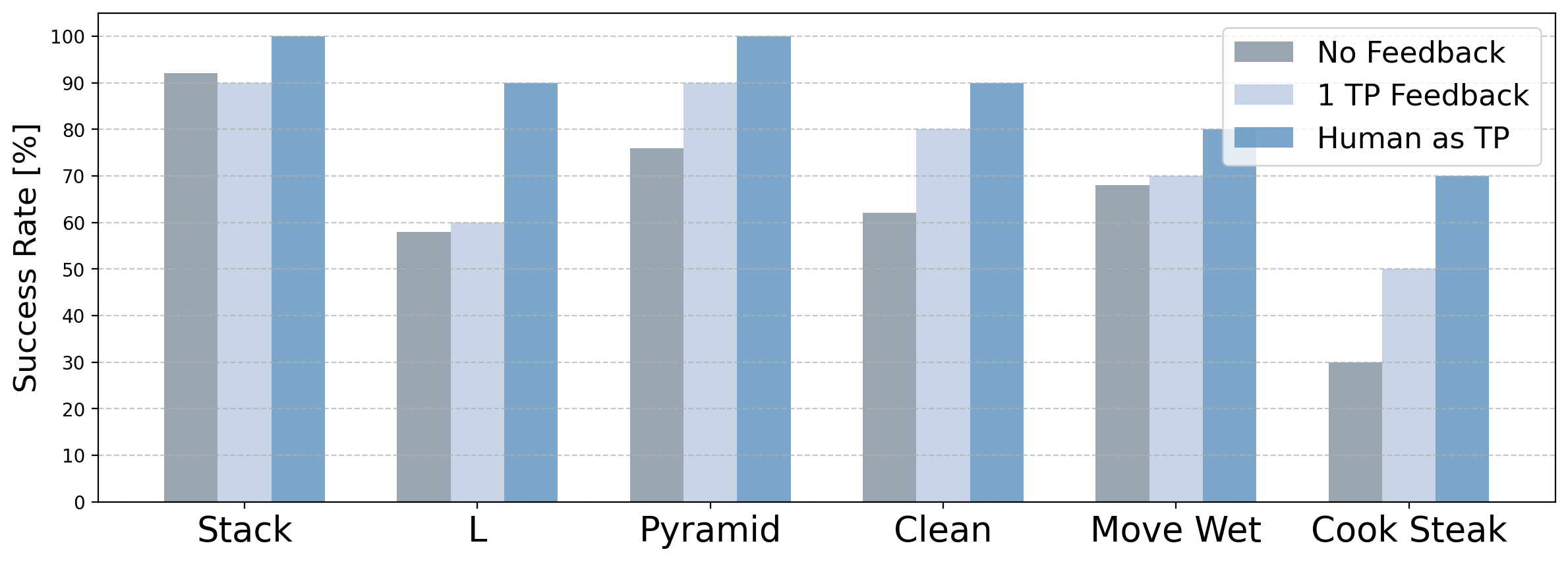}
    \caption{{Task Performance with Feedback}}
    \label{fig:l2ro}
\end{subfigure}
\caption{\textbf{Effects of adding feedback:} (a) Illustrates the interaction between a human and the LLM.  The user asks to double the gripper's speed which leads to the OD multiplying the time by two in the objective function. (b)  Success rate comparison with no human feedback, with one feedback message sent to the TP after the plan is generated, and when an expert human acts as the TP and provides each sub-task instruction to the OD.}
\label{fig:combined_figure}
\end{figure}
We further investigate the ability to integrate feedback into the optimization design. If an action is not carried out correctly, the user can instruct the LLM in natural language on how to readjust for the error. In \cref{fig:l2ro}, we present comparisons of the performance of the LLM at various levels of human intervention for the given tasks. We evaluate the success rate of the system for the six tasks in three different settings: in case of no human intervention, in the setting where the human is allowed to provide one feedback message after the TP has generated a plan, and finally, where the human substitutes the TP.  Success rates, where the human acts as the TP, were evaluated as the average of two different (expert) people carrying out five runs each. 

Since most failures are attributed to an incoherent plan formulated by the TP, the addition of feedback leads to tangible improvements. In fact, the best performance is achieved when the human prompts the \emph{OD} one task at a time. As illustrated in \Cref{fig: OD_feedback}, the intuitiveness and versatility of the designed interface leads to an easy way of incorporating feedback.\looseness=-1

\subsection{Real World Deployment}
We finally evaluate the single-robot tasks (\emph{Stack}, \emph{Pyramid}, \emph{L}, and \emph{Clean Plate}) on a Franka robot, highlighting that our method can directly be deployed on hardware. We again compare the success rates for \emph{VoxPoser}, \emph{CaP}, and \ourmethod as reported in Table \ref{tab:real_world_results}. The experiments confirm the validity of the method in terms of success rate, with an increased collision rate compared to simulation. Often times the collisions are mild, and we attribute them to the imperfect perception system, which has approximately 1cm maximum slack in pose estimation accuracy. 
In \cref{fig:real_world_rollouts}, we additionally demonstrate the versatility and generalizability of our method by deploying it on an additional manipulator (DynaaArm), which differs from the one used in the previous experiments, showcasing that our method can easily be used with different embodiments, thanks to its modular design.
\begin{table}[h!]
	\setlength{\tabcolsep}{5pt}
	\centering
	\begin{tabular}{l|cc|cc}
		\toprule
		\multirow{2}{*}{Method} & \multicolumn{2}{c|}{Stack} &  \multicolumn{2}{c}{L-Shape}  \\ 
		                                  & SR$_\%$$\uparrow$ & Co/Pl/Cd$_\%$$\downarrow$     & SR$_\%$$\uparrow$ & Co/Pl/Cd$_\%$$\downarrow$ \\ \toprule
		CaP \cite{liang2023code}   & \textbf{70}   & 30 /\ \  0 /\ \  0 & 10            & 10 / 40 / 40       \\           
		VoxPoser \cite{huang2023voxposer} & 10            & 90 /\ \  0 /\ \  0    & 20             & 60 /\ \ 0 / 20 \\
		\ourmethod                        & \textbf{70}            & 30 /\ \  0 /\ \  0 & \textbf{70}   &\ \  0 / 30 /\ \  0           \\  \midrule

   & \multicolumn{2}{c|}{Pyramid} &  \multicolumn{2}{c}{Clean Plate}\\ \midrule
		CaP \cite{liang2023code}   &  10            &\ \  0 / 70 / 20   & 70            &\ \  0 / 20 / 10\\ 
		VoxPoser \cite{huang2023voxposer} & 10            & 80 /\ \  0 / 10     & 70            &\ \ 0 /\ \ 0 / 30\\ 
		\ourmethod                          & \textbf{90}   &\ \  0 / 10 /\ \  0   & \textbf{90}            & \ \ 0 / 10 /\ \ 0\\ \bottomrule
	\end{tabular}
	\caption{\textbf{Real World Experiments:} Comparison of \ourmethod against \emph{CaP} and \emph{VoxPoser} on four single-robot tasks in the real world as seen in \Cref{fig: all_tasks}, 10 repetitions per experiment. We report the success rates for each method and the failure reason categorized into Collision (\emph{Co}), Planning (\emph{Pl}) and Code Execution (\emph{Cd}) errors.}
	\label{tab:real_world_results}
\end{table}
\subsection{Limitations}
While \ourmethod achieves impressive results, there still remain challenges that need to be solved. Despite the fact that safety constraints are incorporated directly, the safe execution of all tasks is not guaranteed since there is no verification for the constraints predicted by the LLM. Additionally, especially on real-world deployment, tight integration of visual feedback is valuable to scale the method to diverse environments where pose tracking of objects is challenging or pose information is not sufficient. Finally, dealing with uncertain environments would require the inclusion of (visual) feedback in the Language Module. 

\section{Conclusion and Future Work}
In this work, we propose an architecture to interface natural language instructions with low-level system actions via a language pre-trained module and an optimal control module. By using optimal control, we are able to impose constraints on the actions of the system, which increases its safety and reliability. In this way, we can take advantage of the flexibility of natural language instructions, since we do not need pre-defined functions nor expensive reinforcement learning iterations to ensure good system behavior. 
We illustrate in simulation and on real robots, the effectiveness of our approach as compared to other strategies.
We show that LLMs are suitable for writing control policies as free-form mathematical expressions, which allow for versatile control policies that can solve a set of complex tasks with higher success rates and higher efficiency compared to existing methods. 
For future work, we aim to create guarantees for the safety of both the language module as well as the control module. Additionally, we plan to tightly integrate visual observations and introduce feedback directly to the Language Module.


{\footnotesize
\bibliographystyle{IEEEtran}
\bibliography{IEEEfull,references}
}

\clearpage

\end{document}